\documentclass{article}

\usepackage{PRIMEarxiv}

\usepackage[utf8]{inputenc} % allow utf-8 input
\usepackage[T1]{fontenc}    % use 8-bit T1 fonts
\usepackage{hyperref}       % hyperlinks
\usepackage{url}            % simple URL typesetting
\usepackage{booktabs}       % professional-quality tables
\usepackage{amsfonts}       % blackboard math symbols
\usepackage{nicefrac}       % compact symbols for 1/2, etc.
\usepackage{microtype}      % microtypography
\usepackage{lipsum}
\usepackage{fancyhdr}       % header
\usepackage{graphicx}       % graphics
\graphicspath{{media/}}     % organize your images and other figures under media/ folder

\title{Deepfake tweets automatic detection\thanks{This work was funded by the European Union under the Horizon Europe grant OMINO (grant no 101086321) and by the Polish Ministry of Education and Science within the framework of the program titled International Projects Co-Financed. 
(However, the views and opinions expressed are those of the authors only and do not necessarily reflect those of the European Union or the European Research Executive Agency. Neither the European Union nor the European Research Executive Agency can be held responsible for them.) This research was also carried out with the support of the Faculty of Mathematics and Information Science at Warsaw University of Technology%, %its Laboratory of Bioinformatics and Computational Genomics, 
~and its High-Performance Computing Center.}}

\author{
  Adam Frej, Adrian Kami{\'n}ski \\
  Faculty of Mathematics and Information Science \\
  Warsaw University of Technology \\
  Warsaw\\
  \texttt{\{01151392, 01151387\}@pw.edu.pl} \\
   \And
  Piotr Marciniak, Szymon Szmajdziński \\
  Faculty of Mathematics and Information Science \\
  Warsaw University of Technology \\
  Warsaw\\
  \texttt{\{01151428, 01151438\}@pw.edu.pl} \\
   \And
  Soveatin Kuntur \\
  Faculty of Mathematics and Information Science \\
  Warsaw University of Technology \\
  Warsaw\\
  \texttt{soveatin.kuntur.dokt@pw.edu.pl} \\
   \And
  Anna Wr{\'o}blewska \\
  Faculty of Mathematics and Information Science \\
  Warsaw University of Technology \\
  Warsaw\\
  \texttt{anna.wroblewska1@pw.edu.pl} \\
}

\begin{document}
\maketitle

\begin{abstract}

This study addresses the critical challenge of detecting DeepFake tweets by leveraging advanced natural language processing (NLP) techniques to distinguish between genuine and AI-generated texts. Given the increasing prevalence of misinformation, our research utilizes the TweepFake dataset to train and evaluate various machine learning models. The objective is to identify effective strategies for recognizing DeepFake content, thereby enhancing the integrity of digital communications. By developing reliable methods for detecting AI-generated misinformation, this work contributes to a more trustworthy online information environment.
\end{abstract}

% keywords can be removed

\section{Introduction}
The rise of DeepFake technology in the digital era presents both opportunities and challenges, significantly impacting misinformation through realistic fake content creation, especially in social media tweets~\cite{veerasamy2022rising,pawelec2022deepfakes}. The proliferation of DeepFakes poses a substantial threat to the integrity of information on social media platforms, where the rapid dissemination of false content can lead to widespread misinformation and public distrust. Addressing this issue is critical for maintaining the reliability of digital communications and ensuring that users can distinguish between authentic and manipulated content.

Our study leverages natural language processing (NLP) to develop a DeepFake tweet detection framework, aiming to bolster social media information reliability and pave the way for further research in ensuring digital authenticity. By focusing on the linguistic and contextual nuances that differentiate genuine tweets from AI-generated ones, we seek to create a robust detection mechanism that can be integrated into existing social media platforms to mitigate the spread of misinformation.

Focusing on detecting DeepFake content in tweets, this research employs the TweepFake dataset to evaluate various text representation and preprocessing methods. The TweepFake dataset provides a diverse and comprehensive collection of tweets that facilitate the training and testing of different detection models. We explore effective embeddings and model-generated tweet patterns, utilizing machine learning, deep learning, and transformer technologies. These approaches enable us to capture the subtle differences in writing styles and content structures that are indicative of DeepFake tweets.

The objective is to enhance the detection of GPT2-generated DeepFakes, contributing to the development of robust detection algorithms. By systematically analyzing and comparing the performance of various models, we aim to identify the most effective techniques for distinguishing AI-generated content from genuine tweets. This research not only addresses the immediate challenge of DeepFake detection but also contributes to the broader field of digital content verification, providing insights and methodologies that can be applied to other forms of online misinformation. Ultimately, our work strives to support the creation of a more trustworthy digital information environment, reinforcing the authenticity and reliability of social media communications.

\section{Related Work}
The introduction of the TweepFake dataset has significantly advanced DeepFake detection research, enabling the assessment of different algorithms and highlighting the effectiveness of transformer models like RoBERTa, which achieved an impressive 89.6\% accuracy \cite{tweepfake}. This development underscores a pivotal shift towards sophisticated language models such as BERT, XLNet, and RoBERTa, moving away from traditional bag-of-words approaches. These advanced models have markedly improved textual DeepFake identification by enhancing semantic comprehension and capturing contextual nuances \cite{liu2019roberta, yang2020xlnet}.

In addition to the success of transformer models, novel detection methods like Histogram-of-Likelihood Ranks and energy-based decoding have broadened the scope of detection techniques, though their effectiveness has varied. These innovative approaches have introduced new dimensions to DeepFake detection, focusing on statistical and probabilistic characteristics of text to identify inconsistencies indicative of synthetic content \cite{gehrmann2019gltr}.

Moreover, the research has extended beyond social media tweets to other forms of digital text, such as online reviews, highlighting the widespread challenge of identifying synthetic content across various platforms. Studies focusing on fake reviews have demonstrated the applicability of DeepFake detection techniques to different types of digital content, further illustrating the pervasive nature of this issue and the continuous need for innovative and versatile detection approaches \cite{DBLP:journals/corr/abs-1907-09177}.

Recent advancements in NLP and machine learning have spurred the development of more refined detection algorithms. For instance, the utilization of context-aware embeddings and attention mechanisms in transformer models has significantly improved the accuracy of identifying DeepFake texts. These methods allow models to better understand the intricacies of human language, making it more challenging for AI-generated content to mimic genuine text convincingly.

Furthermore, interdisciplinary approaches that combine linguistic analysis with computational techniques have shown promise in enhancing DeepFake detection capabilities. By integrating insights from cognitive science, psycholinguistics, and computational linguistics, researchers are developing more robust frameworks that can detect subtle anomalies in AI-generated texts.

The ongoing evolution of detection strategies, coupled with the growing sophistication of DeepFake generation techniques, underscores the importance of continuous innovation in this field. As AI-generated content becomes increasingly indistinguishable from genuine text, the development of advanced detection methodologies will be crucial in maintaining the integrity of digital communications. The collective efforts in this area aim to create a more secure online environment, ensuring that users can trust the authenticity of the information they encounter.

\section{Datasets} \label{sec:data}
Our research utilized the TweepFake dataset \cite{tweepfake} and GPT-2 generated texts \cite{ippolito-etal-2020-automatic}, focusing on DeepFake detection.

\textbf{TweepFake Dataset} \cite{tweepfake} contains 25,572 tweets, split evenly between real tweets from 17 Twitter accounts and fake tweets from 23 bot accounts. This dataset provides a diverse set of fake tweet samples for analysis, encompassing various styles and topics. The balanced distribution of real and fake tweets ensures a comprehensive training and evaluation environment for our models, allowing them to learn the distinguishing features of DeepFake content effectively.

\textbf{GPT-2 Generated Datasets} \cite{ippolito-etal-2020-automatic} comprises texts created by GPT-2 \cite{radford_language_2019}, including 500,000 training samples and 5,000 samples each for validation and testing, evenly split between human and machine-generated texts. This extensive dataset is instrumental in training our models, providing a rich source of diverse and challenging examples that span a wide range of topics and writing styles. The inclusion of both training and evaluation samples ensures that our models are rigorously tested and validated, enhancing their ability to generalize to new, unseen DeepFake content.

These datasets collectively support the training and testing of our models, facilitating a comprehensive evaluation of our detection techniques against varied DeepFake content. By leveraging these datasets, we can benchmark our approaches against established baselines and assess their effectiveness in real-world scenarios. The combination of the TweepFake dataset's focus on social media tweets and the GPT-2 generated texts' broader scope of synthetic content allows for a robust examination of our models' capabilities in different contexts.

Furthermore, the diversity within these datasets helps in understanding the nuances and challenges associated with detecting DeepFake content. The TweepFake dataset's real-world relevance, combined with the controlled generation of texts in the GPT-2 dataset, provides a well-rounded foundation for developing and testing advanced DeepFake detection methodologies. Through detailed analysis and iterative refinement of our models using these datasets, we aim to contribute significantly to the field of digital content verification and the broader effort to combat misinformation.

\section{Our Approach}
\label{sec:approach}

Our study leveraged the TweepFake dataset and GPT-2 generated texts, adhering to an 80\%-10\%-10\% split for training, validation, and testing, aiming for direct comparison with prior work \cite{tweepfake}. Our approach focused on several key areas:

\begin{enumerate}
    \item \textbf{Identifying Effective Text Representations and Preprocessing for TweepFake Dataset DeepFake Detection}: We investigated various methods for text representation and preprocessing to enhance the accuracy of DeepFake detection. This included examining different tokenization strategies, stop word removal techniques, and the application of stemming or lemmatization to normalize the text data.
    
    \item \textbf{Evaluating Machine Learning, Deep Learning, and Transformer Model Efficacy in DeepFake Detection}: Our evaluation covered a broad spectrum of models to determine their effectiveness in identifying DeepFake tweets. This included traditional machine learning models, state-of-the-art deep learning architectures, and advanced transformer models, providing a comprehensive analysis of their performance.
    
    \item \textbf{Assessing Emoticons, Mentions, Misspells, and URLs' Roles in Distinguishing Between Human and Automated Tweets}: We analyzed the impact of specific tweet characteristics, such as emoticons, mentions, misspellings, and URLs, on the ability to distinguish between human-generated and AI-generated tweets. This aspect of the study aimed to identify unique features that could improve detection accuracy.
    
    \item \textbf{Enhancing the Detection Algorithm with Advanced Algorithm-Generated DeepFakes}: We aimed to enhance the robustness of our detection algorithms by incorporating advanced algorithm-generated DeepFakes into our training and evaluation processes. This approach helped ensure that our models remained effective against increasingly sophisticated DeepFake techniques.
\end{enumerate}

To achieve these objectives, we explored various embeddings and bot-generated tweet patterns, utilizing tools such as NLTK \cite{BirdNLTK} and SpacyTextBlob \cite{spacy} for data insights. Our preprocessing pipeline involved several key steps:

\begin{itemize}
    \item \textbf{Tokenization}: Breaking down text into individual tokens to facilitate analysis.
    \item \textbf{Stop Word Removal}: Eliminating common but uninformative words to reduce noise in the data.
    \item \textbf{Stemming or Lemmatization}: Normalizing words to their base or root forms to improve consistency in text representation.
\end{itemize}

Our analysis spanned multiple machine learning models, including LightGBM \cite{ke2017lightgbm}, XGBoost \cite{Chen2016XGBoost}, Random Forest \cite{ho1995random}, Logistic Regression \cite{cox1958regression}, and SVM \cite{cortes1995support}, using TF-IDF and BERT embeddings \cite{devlin2019bert}. In addition to traditional machine learning approaches, we employed deep learning networks such as Convolutional Neural Networks (CNN), Gated Recurrent Units (GRU), and hybrid CNN+GRU architectures.

Furthermore, we investigated the performance of transformer models, including xlm-roberta-base \cite{xlm-roberta}, distilbert-base-uncased \cite{sanh2020distilbert}, and GPT-2 \cite{radford_language_2019}. These models were evaluated based on key metrics such as accuracy, precision, recall, and F1 score to refine our Twitter DeepFake detection techniques.

Our comprehensive approach allowed us to identify the most effective strategies for detecting DeepFake tweets, contributing to the development of robust and reliable detection algorithms that can enhance the integrity of digital communications.

\section{Results}\label{sec:results}

Table~\ref{tab:res_top10}, catalogs the premier models by their balanced accuracy. Remarkably, the ROBERTA model, leveraging the unaltered TweepFake dataset (raw data in its original, unmodified state), emerged as superior in both balanced accuracy and F1 score. An evaluation across data processing techniques—specifically, unprocessed (raw) data, BERT embeddings (where text is converted into vectors capturing nuanced contextual relationships through the BERT algorithm), and lemmatization (simplifying words to their base form for uniformity)—demonstrated that raw data and BERT embeddings are particularly potent for deepfake detection.

Table~\ref{tab:res_grouped_top10} delves into the models' performance against different tweet sources, illustrating the nuanced challenge of identifying GPT2-engineered deepfakes. It becomes evident that RNN-generated tweets are more straightforward to detect, contrasting with the subtlety of GPT2-produced forgeries, which signifies their advanced deceptive quality.

\begin{figure*}[h]
    \centering
    \small
    \includegraphics[width=\linewidth]{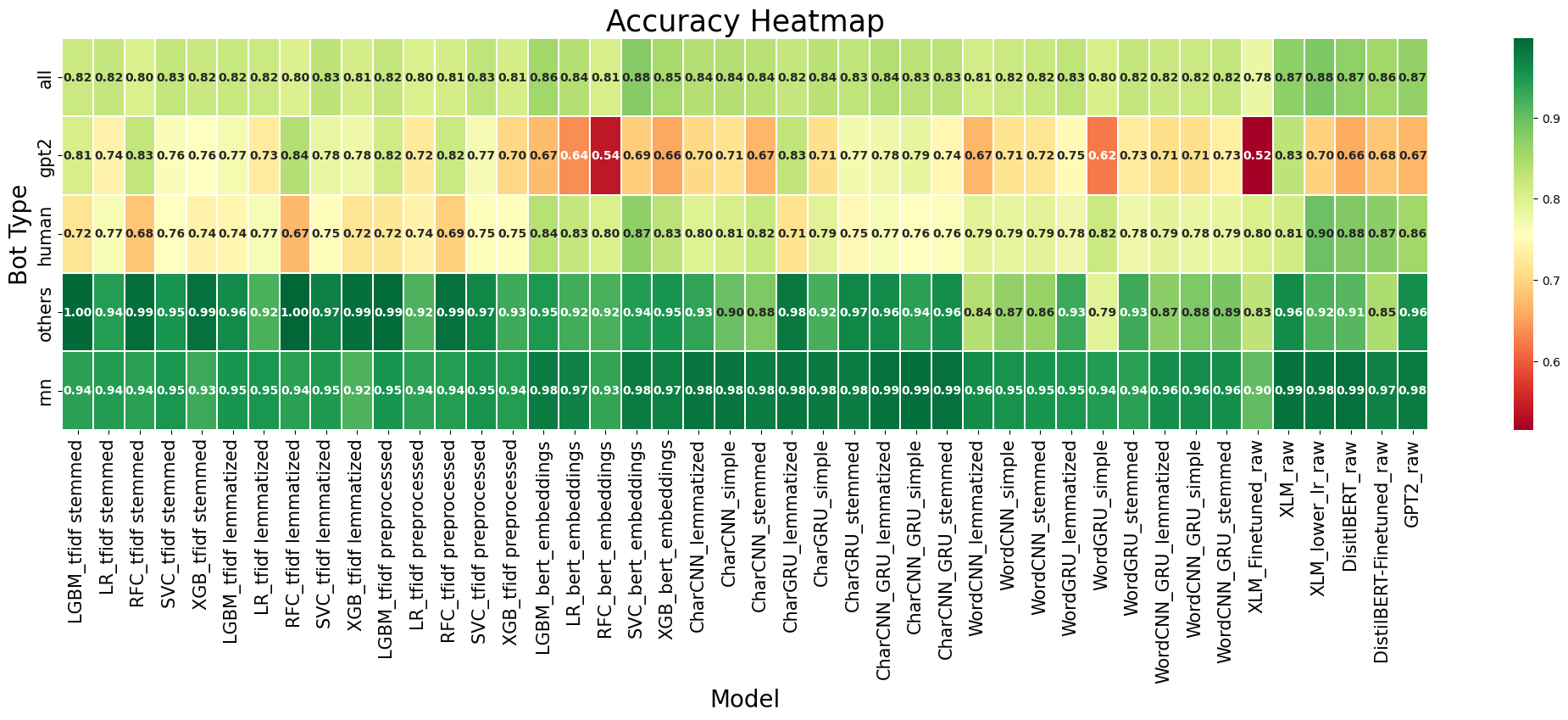}
    \caption{Accuracy for class type and every model}
    \label{fig:acc_heatmap}
\end{figure*}

Our investigations have yielded significant progress in the detection of deepfakes, especially those generated by GPT2, establishing new benchmarks in this complex arena. This advancement is crucial, considering the sophisticated nature of such fabrications. Figure~\ref{fig:acc_heatmap} offers a visual exposition of the detection challenge posed by GPT2-generated deepfakes and the relative proficiency of transformer-based models in this domain.
\begin{table*}[htbp]
    \caption{Top 10 DeepFake Detection Models Ranked by Accuracy}
    \centering
    \small
    \begin{tabular}{rrrrll}
    \toprule
    model & pre-processing & ba & f1 & precision & recall \\
    \midrule
    \textit{ROBERTA (TweepFake)} & \textit{raw} & \textit{\textbf{0.896}} & \textit{\textbf{0.897}} & \textit{0.891} & \textit{0.902} \\
    XLM2 & raw & 0.8835 & 0.8821 & \textbf{0.8934} & 0.8711 \\
    SVC & bert & 0.8757 & 0.8763 & 0.8729 & 0.8797 \\
    XLM1 & raw & 0.8713 & 0.8786 & 0.8328 & \textbf{0.9297} \\
    DISTIL\_BERT0 & raw & 0.8698 & 0.8686 & 0.8773 & 0.8602 \\
    GPT2 & raw & 0.8671 & 0.8686 & 0.8593 & 0.8781 \\
    LGBM & bert & 0.8561 & 0.8590 & 0.8429 & 0.8758 \\
    DISTIL\_BERT1 & raw & 0.8554 & 0.8529 & 0.8681 & 0.8383 \\
    XGB & bert & 0.8518 & 0.8546 & 0.8395 & 0.8703 \\
    CharCNN+GRU & lemmatized & 0.8408 & 0.8518 & 0.7975 & 0.9141 \\
    LR & bert & 0.8393 & 0.8416 & 0.8304 & 0.8531 \\
    \bottomrule
    \end{tabular}
    \label{tab:res_top10}
\end{table*}
\begin{table*}[htbp]
    \caption{Model Accuracy Across Different Sources of DeepFakes}
    \small
    \label{tab:res_grouped_top10}
\small
    \centering
\begin{tabular}{lrrrrr}
    \toprule
    & \multicolumn{5}{c}{\textbf{TWEET CREATOR (CATEGORY)}} \\ \cmidrule{2-6}
    model name & ALL & GPT2 & HUMAN & OTHERS & RNN \\
    \midrule
    \textit{ROBERTA\_FT (TweepFake)} & \textit{\textbf{0.896}} & \textit{0.74} & \textit{0.89} & \textit{0.95} & \textit{\textbf{1.00}} \\
    \textbf{XLM2\_raw} & 0.8835 & 0.6953 & \textbf{0.8959} & 0.9153 & 0.9830 \\
    \textbf{SVC\_bert\_embeddings} & 0.8757 & 0.6927 & 0.8717 & 0.9442 & 0.9782 \\
    \textbf{XLM1\_raw} & 0.8714 & \textbf{0.8307} & 0.8130 & 0.9607 & 0.9854 \\
    \textbf{DisitlBERT0\_raw} & 0.8698 & 0.6589 & 0.8795 & 0.9112 & 0.9879 \\
    \textbf{GPT2\_raw} & 0.8671 & 0.6693 & 0.8560 & 0.9587 & 0.9782 \\
    \textbf{LGBM\_bert\_embeddings} & 0.8561 & 0.6745 & 0.8365 & 0.9483 & 0.9782 \\
    \textbf{DistilBERT1\_raw} & 0.8554 & 0.6849 & 0.8725 & 0.8471 & 0.9709 \\
    \textbf{XGB\_bert\_embeddings} & 0.8518 & 0.6562 & 0.8333 & 0.9525 & 0.9733 \\
    \textbf{CharCNN\_GRU\_lemmatized} & 0.8409 & 0.7760 & 0.7676 & \textbf{0.9628} & 0.9854 \\
    \textbf{LR\_bert\_embeddings} & 0.8393 & 0.6380 & 0.8255 & 0.9236 & 0.9709 \\
    \bottomrule
\end{tabular}
\end{table*}

\section{Conclusion}
Our study has demonstrated that current models can accurately detect DeepFakes generated by GPT-2, highlighting the effectiveness of advanced natural language processing techniques in addressing this challenge. However, it is essential to recognize the rapid advancements in language models, which continually push the boundaries of how we understand and generate text. While GPT-2 represented a significant leap forward, the emergence of newer and more complex models, such as GPT-3 and beyond, fundamentally alters the landscape of language generation and detection.

This dynamic environment underscores the necessity for detection methods that evolve in tandem with these advancements. As language models become increasingly sophisticated, distinguishing between authentic and synthetic content will become more challenging. This reality emphasizes the critical need for ongoing updates and enhancements in detection methodologies to keep pace with the evolving capabilities of AI-generated content.

Future research must focus not only on refining existing detection techniques but also on anticipating and adapting to the advancements in language model capabilities. Developing more robust and adaptive algorithms will be crucial in maintaining the efficacy of DeepFake detection. This approach is vital for safeguarding the integrity of online communication, ensuring that users can trust the authenticity of the information they encounter.

Moreover, interdisciplinary collaboration will play a pivotal role in advancing DeepFake detection research. Integrating insights from fields such as cognitive science, computational linguistics, and cybersecurity can lead to the development of more comprehensive and resilient detection frameworks. By fostering a multidisciplinary approach, we can better prepare for the challenges posed by the continual evolution of DeepFake technology.

In conclusion, our study highlights both the current successes and the future challenges in DeepFake detection. The rapid pace of language model innovation necessitates a proactive and forward-looking approach to detection research. By continually updating and enhancing our methods, we can better protect the integrity of digital communications and mitigate the impact of increasingly sophisticated DeepFake content. This ongoing effort is essential for maintaining trust and authenticity in the digital age.
%Bibliography
\bibliographystyle{plain}  
\bibliography{references}

\end{document}